\documentclass[preprint,11pt]{elsarticle}

\usepackage{amssymb}
\usepackage{amsmath}
\usepackage{caption}  
\usepackage{subcaption}
\usepackage{booktabs}
\usepackage{url}
\usepackage{float}

\usepackage[
  top=1in,
  bottom=1in,
  left=1in,
  right=1in
]{geometry}

\journal{ArXiv}

\begin{document}

\begin{frontmatter}



\title{LLMs can persuade only psychologically susceptible humans on societal issues, via trust in AI and emotional appeals, amid logical fallacies} %% Article title


% \author{} %% Author name

% \affiliation{organization={},%Department and Organization

% ==============================================================
% ==============================================================

\author[1]{Alexis Carrillo}
\author[1]{Salvatore Citraro}
\author[1]{Ali Aghazhadeh Ardebili}
\author[1]{Enrique Taietta}
\author[2]{Giulio Rossetti}
\author[3]{Emilio Ferrara}
\author[4,5]{Giuseppe Alessandro Veltri}
\author[1]{Massimo Stella\corref{cor1}}

\affiliation[1]{organization={CogNosco Lab, Department of Psychology and Cognitive Science, University of Trento},
            city={Rovereto},
            country={Italy}}

\affiliation[2]{organization={Institute of Information Science and Technologies ``Alessandro Faedo", National Research Council},
            % city={},
            country={Italy}}

\affiliation[3]{organization={Thomas Lord Department of Computer Science, University of Southern California},
            city={Los Angeles},
            state={California},
            country={USA}}

\affiliation[4]{organization={Department of Sociology and Social Research, University of Trento},
            city={Trento},
            country={Italy}}
            
\affiliation[5]{organization={Centre for Behavioural and Implementation Sciences in Medicine (BISI), National University of Singapore},
            city={Singapore},
            country={Singapore}}

\cortext[cor1]{Corresponding author. Email: massimo.stella-1@unitn.it}

% ==============================================================
% ==============================================================

%% Abstract
\begin{abstract}
Scarce longitudinal evidence examines LLMs' persuasiveness and humanness along time-evolving psychological frameworks. We introduce Talk2AI, a longitudinal framework quantifying psycho-social, reasoning and affective dimensions of LLMs' persuasiveness about polarizing societal topics. In a four-way longitudinal setup, Talk2AI's 770 participants engaged in structured conversations with one of four leading LLMs on topics like climate change, social media misinformation, and math anxiety. This produced 3,080 conversations over 60,000 turns. After each wave, participants reported conviction in their initial topic stance, perceived opinion change, LLM's perceived humanness, a self-donation to the topic and a textual explanation. Feedback time series showed longitudinal inertia in convictions, indicating some human anchoring to initial opinions even after repeated exposure to AI-generated arguments. Interestingly, NLP analyses revealed that both humans and LLMs relied on fallacious reasoning in 1 conversational quip every 6, countering the ``LLMs as superior systems" stereotype behind LLMs' cognitive surrender. LLMs' perceived humanness was most learnable from sociodemographic, psychological and engagement features ($R^2=0.44$), followed by opinion change ($R^2=0.34$), conviction ($R^2=0.26$) and personal endowment ($R^2=0.24$). Crucially, explainable AI (XAI) indicated: (i) the presence of individuals more susceptible to LLM-based opinion changes; (ii) psychological susceptibility to LLM-convincing consisted of having more trust in LLMs, being more agreeable and extraverted and with a higher need for cognition. A multiverse approach with mixed-effects models confirmed XAI results, alongside strong individual differences. Talk2AI provides a grounded framework and evidence for detecting how GenAI can influence human opinions via multiple psycho-social pathways in AI-human digital platforms.
\end{abstract}


% \begin{highlights}
% \end{highlights}

%% Keywords

\end{frontmatter}



\section{Introduction}
\label{sec1}
%% Labels are used to cross-reference an item using \ref command.

Large Language Models (LLMs) are now used routinely as conversational tools. People turn to them to obtain information, summarize news, clarify scientific concepts, and discuss personal or political questions. Survey evidence indicates that roughly 1 in 3 U.S. adults report having used ChatGPT, and adoption exceeds 50\% among adults under 30 \cite{pew2025chatgpt,chatterji2025people}. At the same time, the footprint of generative AI continues to expand: Industry estimates suggest that it could add between 2.6 and 4.4 trillion U.S. dollars per year in value across major sectors \cite{singla2025state}. Many of these conversations concern sensitive topics such as climate change \cite{nguyen2025misrepresentation,ferrara2025information}, misinformation \cite{ferrara2025information,hackenburg2025levers}, and Education \cite{poquet2023chatgpt,ciringione2025math}. Yet, there is still little longitudinal evidence on how repeated interactions with conversational AI shape users' beliefs, metacognitive judgments, and cognitive organization over time \cite{matz2024potential,carrasco2024large}.

Controlled experiments show that short interactions with AI-generated content can measurably influence attitudes and preferences \cite{bai2025llm}. For example, a preregistered survey experiment with 8,221 respondents found that GPT-3--generated propaganda articles produced agreement shifts comparable to those produced by real-world propaganda texts \cite{goldstein2024persuasive}. In conversational settings, a randomized debate experiment with 820 participants showed that GPT-4, when given access to basic demographic information about the user, increased the odds of opinion change by 81.7\% relative to human persuaders \cite{salvi2025conversational}. Other work shows that AI systems can improve tone and reciprocity in divisive discussions \cite{argyle2023leveraging} and that personalized messages generated by LLMs can achieve modest but reliable persuasive advantages over non-targeted communication \cite{simchon2024persuasive}. Large-scale studies further indicate that conversational systems can steer attitudes across a wide range of political topics even with minimal user information \cite{hackenburg2025levers}. Together, these results suggest that LLM-generated text and dialogue can produce statistically detectable shifts in attitudes and preferences, raising concerns about scalable persuasion and microtargeting in digital communication \cite{rossetti2024social,argyle2025testing}.

However, the current evidence base leaves a critical gap. Most studies operationalize persuasion as a single end-point---for example, immediate attitude change, stated preferences, or voting intentions---rather than as a multidimensional process \cite{carrasco2024large}. They therefore leave three issues unresolved. First, most studies examine short, one-shot exposures rather than repeated conversations across time \cite{bai2025llm,goldstein2024persuasive}. Second, even interactive designs often ask participants to debate pre-specified pro/con claims that may not map onto their own concerns or lived positions \cite{salvi2025conversational}. Third, current work rarely examines the logical fallacies that humans and LLMs may share in dialogue \cite{hackenburg2025levers,salvi2025conversational}. While pioneering and innovative, these studies also provide limited leverage for disentangling the semantic, structural, emotional, and social features that make LLMs persuasive \cite{simchon2024persuasive,argyle2023leveraging,hackenburg2025levers}. Recent syntheses help sharpen this gap. AI is increasingly conceptualized not merely as a delivery channel but as a persuasion agent whose effects depend jointly on properties of the system, characteristics of the recipient, and the context of interaction \cite{watson2024ai}. Meta-analytic evidence further indicates that AI communicators are, on average, about as persuasive as human communicators overall, but comparatively weaker at shifting behavioral intentions than perceptions and attitudes \cite{huang2023persuasive}. This distinction is crucial for longitudinal human--LLM dialogue because repeated conversations may alter social appraisal, felt influence, certainty, and downstream action to different extents rather than producing a single undifferentiated persuasion effect.

Importantly, the same interaction can shift one component of persuasion without shifting others \cite{brady2025dual,blankenship2023certainty}. Users may judge an argument as compelling while remaining anchored to their initial stance, or they may report influence without translating it into action. These dissociations matter because they point to distinct psychological mechanisms and different degrees of temporal stability. Dual-process theories such as the Elaboration Likelihood Model (ELM) and the Heuristic-Systematic Model (HSM) provide a natural framework for explaining them \cite{chaiken1989heuristic,petty2009elaboration}. Conversational LLMs are well positioned to activate both routes. On the more effortful route, they can produce reasons, counterarguments, and explanations that invite scrutiny \cite{ferrara2025information}. On the lower-elaboration route, they provide fluent language, personalization, rapid contingency, confident tone, and socially appropriate turn-taking---all of which can function as cues for credibility, expertise, and benevolence \cite{brady2025dual,metzger2013credibility}. Recent AI scholarship sharpens this dual-process view rather than replacing it: The persuasive effects of AI depend on how message quality, system cues, and user dispositions combine in context \cite{watson2024ai}. Meta-analytic evidence likewise suggests that AI communicators are roughly as persuasive as human communicators overall, but comparatively weaker at changing behavioral intentions than perceptions and attitudes \cite{huang2023persuasive}. In repeated dialogue, prior trust and prior reliance can shape subsequent trust and reliance toward AI advice, making interaction history a central moderator of persuasive effects \cite{kahr2024trust}. Need for Cognition is therefore a relevant moderator \cite{cacioppo1982need}: Individuals who enjoy effortful thinking are more likely to scrutinize arguments, whereas others may rely more heavily on heuristics under low involvement or cognitive strain \cite{petty2009elaboration,chaiken1989heuristic}. Trust in the source is another key moderator because it changes the threshold for acceptance and the degree of scrutiny \cite{metzger2013credibility}. For LLMs, trust is likely shaped by prior experience, perceived competence, expectations of truthfulness, and the interactional history between user and system \cite{carrasco2024large,kahr2024trust}.

Dual-process theories explain how messages are processed, but conversational AI adds a second layer: users often respond to interactive systems as social partners. The Computers as Social Actors (CASA) framework \cite{gambino2020building,xu2022deep} predicts that when a system displays social cues---language, politeness, empathy, and contingency---people import social scripts from human interaction and form social inferences about the system. These inferences are often experienced as ``humanness'' or social presence. Recent meta-analytic evidence on text-based conversational agents supports this interpretation: human-like social cues produce a small but reliable increase in users' social responses, with stronger effects on perceptions, affect, rapport, and trust than on overt behavior \cite{klein2025socialcues}. In persuasion terms, perceived humanness can therefore act as a potent heuristic cue. It can increase perceived warmth and credibility, reduce defensive processing, and shift how users interpret the same content. A framework that ignores CASA risks misattributing persuasion to semantic content alone, when part of the effect may instead arise from social perception and trust \cite{gambino2020building,de2025measuring}.

Current AI literature also lacks psychologically interpretable process measures that connect dialogue to persuasion. Many studies quantify outcomes but do not trace mechanisms within the interaction itself \cite{simchon2024persuasive,argyle2023leveraging,hackenburg2025levers}. Two process measures are especially relevant for LLM dialogue. The first is emotional language. Emotion can increase attention, shape appraisal, and signal value or urgency. It can therefore amplify persuasion through both ELM/HSM routes: as substantive content when users elaborate, and as an affective cue when they do not \cite{chaiken1989heuristic,petty2009elaboration}. The second is fallacious reasoning \cite{cau2025selective,jin2022logical}. Fallacies are not merely errors. They can serve social goals, such as saving face, and cognitive goals, such as reducing complexity. Recent cognitive work shows that susceptibility to poor arguments depends jointly on prior attitudes and cognitive sophistication \cite{marin2024poorarguments}. In real conversations, both humans and LLMs may therefore produce arguments that sound plausible while remaining logically weak \cite{brady2025dual,carrasco2024large}. Measuring fallacy patterns provides a concrete, interpretable link between conversational form and persuasive outcomes.

These gaps matter beyond academia. If LLMs can influence judgments quickly, repeatedly, and at scale, the risk is not limited to overt attitude change. Key concerns are \emph{cognitive surrender} \cite{shaw2026thinking} and \emph{cognitive sovereignity} \cite{branda2026comfort}, where users may defer to an AI's fluent output with minimal scrutiny, internalizing it as third reasoning system rather than as mere external advice. This instance of cognitive offloading \cite{risko2016cognitive} might be due to several reasons, including a perception of ``LLMs as superior systems" capable of producing factual, unbiased knowledge \cite{shaw2026thinking,breum2024persuasive}, thus worthy of trust \cite{de2025measuring}. Extensive evidence shows that LLMs can produce biased knowledge \cite{williams2026biased,cau2025selective,williams2026biased}, instead. Furthermore, state-of-the-art LLMs can even recognize their own biases when prompted with “could you be wrong” \cite{hills2026could}. Recent preregistered experiments show that participants consult the AI frequently and follow it even when it is wrong \cite{shaw2026thinking} or biased \cite{williams2026biased}. Analogously, Breum and colleagues showed that human raters indicated as more convincing those LLM-based arguments with more factual knowledge, resembling more human expert-level feedback \cite{breum2024persuasive}. These patterns are concerning for several reasons. Because generative models also lower the cost of personalized persuasion \cite{chui2023economic}, they can amplify social-media polarization \cite{stella2018bots}, synthetic-media flooding \cite{ferrara2016rise}, and disinformation \cite{adam2023artificial,gavriil2025fog}. For regulation and safeguards---including under the EU AI Act---we therefore need longitudinal, multidimensional, and psychologically interpretable measures of persuasive influence \cite{gavriil2025fog}. This is where Talk2AI comes into play.

\subsection{Filled Research Gaps and Research Questions}

To fill the above gaps, we introduce Talk2AI: A data-informed, interpretable, and psychologically grounded framework. As highlighted in Figure \ref{fig:1}, Talk2AI provides a longitudinal test bed for studying how conversations with LLMs shape human reasoning on socially polarizing topics such as climate change, misinformation, and math anxiety. Using longitudinal data from more than 3,000 human--LLM conversations across four weekly waves, Talk2AI examines whether, when, and through which mechanisms LLMs can persuade specific classes of users, including emotional storytelling, logical reasoning, and heuristics appealing to users' personality traits, cognitive styles and sociodemographic backgrounds. Talk2AI is built around a clear theoretical commitment: To provide a rigorous foundation for understanding how conversational AI systems influence human reasoning and perceptions in everyday societal contexts.

\subsection{Scope and Data Structure}

As reported in Figure \ref{fig:1}, Talk2AI's participants ($N = 770$) engaged in 4 synchronous chat sessions about a given topic with an assigned LLM assistant, cf. \textit{Materials and Methods, Study Design}. Sessions were separated by a one-week interval. Each conversation followed a standardized framework consisting of 10 user's inputs and 10 LLM's responses. In total, 3,080 conversations (thus 61,600 quips) were collected and structured across multiple descriptors combining participants' background and their psychological proxies, along with textual features derived from the interactions. The user level included detailed sociodemographic profiles and responses to validated psychometric instruments, cf. \textit{Materials and Methods, User Level Features} and \textit{Time-Invariant Psychological Features}. The post-interaction data contained the post-session feedback metrics, collected via a response slider presented to participants. These feedbacks quantify users' conviction stability ($Q0$), their self-reported opinion change on the discussed topic ($Q1$), as well as LLMs' perceived humanness ($Q2$), users' personal endowment or self-donation through a donation game ($Q3$) and a final textual feedback about the quality of the interaction ($Q4$), cf. \textit{Materials and Methods, Feedback Metrics}.
The linguistic level transforms participant-assistant exchanges into features proxying engagement, the emotional tone of the conversation, and eventual persuasive strategies present in both users and LLMs (cf. \textit{Materials and Methods, Engagement, Emotion, and Fallacy Detection}).
In the following, we leverage this integrated structure to analyze the dynamics of feedback metrics and the trajectories of persuasive responses through fallacy patterns. Finally, we investigate the drivers of feedback responses.

\section{Materials and Methods}

\subsection{Study Design}

Recruitment was conducted via a certified online panel managed by Bilendi Italia (https://www.bilendi.it/ - Accessed: 23/03/2026). The initial invitation targeted a sample of 2,545 individuals representative of the adult Italian population aged 18 to 69 years. Data collection took place during spring 2025. From the initial cohort, 814 participants completed all four conversational waves. Following rigorous data quality control procedures, the final dataset comprises $N=770$ fully profiled users.
The study employed a within-subject longitudinal design consisting of 4 distinct waves with consecutive sessions separated by a one-week interval. During each session, participants engaged in a structured dialogue with an AI assistant limited to ten conversational turns per session. Participants were randomly assigned to interact with one of four state-of-the-art Large Language Models: GPT-4o \cite{hurst2024gpt}, Claude Sonnet 3.7 (https://www.anthropic.com/news/claude-3-7-sonnet - Accessed: 23/03/2026), DeepSeek-chat V3 \cite{liu2024deepseek}, or Mistral 8b (https://docs.mistral.ai/models/ministral-8b-24-1 - Accessed: 23/03/2026). Each user was assigned a specific discussion topic, constant across all four waves, among climate change, math anxiety, or health misinformation. 
Data collection was administered via a custom web platform. The procedure followed a two-phase structure consisting of an initial registration to gather stable user traits followed by four longitudinal sessions where psychometric profiling, conversational interaction, and feedback were repeated to track temporal dynamics.

\subsection{User Level Features}

During the initial registration phase, participants completed a survey to establish their sociodemographic profile. Self-reported gender, date of birth for age calculation, and total household size were recorded. Education level was classified according to the European Qualifications Framework on a scale from 1 (Primary School) to 8 (Doctoral Degree). Employment status categorized users as self-employed, employee, gig-worker, or unemployed. AI literacy was assessed via a binary indicator regarding prior familiarity with language models like ChatGPT. Socioeconomic standing was measured using both subjective and objective indicators. Subjective status utilized an adaptation of the MacArthur Scale of Subjective Social Status \cite{Adler2000subjective} where participants placed themselves on a visual ladder representing society's resource distribution. Objective financial distress was assessed by asking participants if they had experienced difficulty paying bills in the last 12 months.

Prior to each interaction, participants completed a battery of psychometric scales rated on a 5-point Likert scale. General trust and attitude toward AI assistants were measured using the 8-item Trust in Large Language Models Inventory (TILLMI, \cite{de2025measuring}). Cognitive disposition was assessed using the 6-item short-form Need for Cognition scale \cite{cacioppo1982need}. Personality traits were captured using the 10-item short version of the Big Five Inventory covering Extraversion, Agreeableness, Conscientiousness, Neuroticism, and Openness \cite{guido2015italian}. % Factor analysis of this scale identified two sub-dimensions: NFC-Seek which measures the intrinsic motivation to engage in complex cognitive tasks, and NFC-Diligence which measures persistence in behavioral discipline regarding mental tasks. Personality traits were captured using the 10-item short version of the Big Five Inventory covering Extraversion, Agreeableness, Conscientiousness, Neuroticism, and Openness.
Cf. \textit{SI, Feature Descriptions} for the complete questions presented to participants.

\subsection{Conversation and Feedback Scores}

After the provisioning of psychometric data, users engaged in a conversation with a given LLM assistant, designed along psychological principles of metacognition (e.g. reflecting on one's reasoning flaws \cite{hills2026could}). Every user was attributed a topic, known to them, and an LLM assistant, unknown to them. They were asked to start by writing at least 50 words about their initial stance to a given topic. The LLM then started replying to the human. The user was then nudged to identify mistakes in the LLM's arguments via pre-compiled options. These options included: (i) Please give me more details about..., (ii) What you wrote is incorrect because..., and (iii) Are you sure about what you wrote? All options were directed from the user towards the LLM and they appeared only in the first 3 rounds of conversation. The user was always given the option to write their own reply. The conversation lasted for 10 rounds.

Upon concluding the conversation, participants provided structured feedback to evaluate the interaction. Conviction stability ($Q0$) was measured by asking users to rate the strength of their initial arguments on a scale of 1 to 100. Self-reported opinion change ($Q1$) was assessed by asking participants to quantify how much the conversation altered their views on the assigned topic. Perceived humanness ($Q2$) was evaluated by asking users to rate how much the interaction resembled a dialogue with a human. Self-donation or personal endowment ($Q3$) was operationalized using a Dictator Game proxy where participants engaged in a hypothetical scenario involving the allocation of 100 Euros between themselves and a relevant charity. The score registered in the response indicates the amount of the 100 euros to donate that users chooses to keep for themselves. Finally, qualitative feedback ($Q4$) was gathered via an open-ended text field requiring a minimum of 50 words regarding users' current thoughts on the topic.

\subsection{Engagement, Emotion and Fallacy Detection}
Textual features were extracted to measure: (1) participants' engagement \cite{jarvis2013capturing}, (2) the emotional tone of the conversations \cite{semeraro2025emoatlas}, (3) and eventual logical fallacies \cite{jin2022logical} in participant and LLM arguments. First, we extracted simple textual features at the start and the end of the experiments, namely from the user's first quip of the first wave and from the final textual feedback $Q4$. The features are the total number of tokens and the lexical diversity \cite{jarvis2013capturing} measured as the type–token ratio, i.e., the proportion of unique lemmas relative to total tokens.

Statistically significant emotional scores for eight basic emotions (Anger, Anticipation, Disgust, Fear, Joy, Sadness, Surprise, and Trust) were extracted to assess the emotional tone of the conversations. We used the EmoAtlas library \cite{semeraro2025emoatlas}, separately for user and assistant texts in each session, e.g., the feature \textit{LLM Ang.(z)} represents the z-score that quantifies the presence of anger in the LLM assistant at the wave level, thus aggregating individual quips. Emotion scores were also computed for the first wave and the final textual feedback to provide additional engagement-related features for the user.

We quantified logical fallacies for users' and LLMs' texts separately using a fine-tuned DistilBERT model\footnote{\url{https://huggingface.co/q3fer/distilbert-base-fallacy-classification}} trained on a dataset of common fallacy examples grouped into 13 categories \cite{jin2022logical}, cf. Fig. \ref{fig:fallacies} legend for the complete list. %: \textit{Ad Hominem} (AdH), \textit{Ad Populum} (AdP), \textit{Credibility} (C), \textit{Circular Reasoning} (CR), \textit{Appeal to Emotion} (Em), \textit{Equivocation} (Eq), \textit{Extension} (Ex), \textit{False Causality} (FC), \textit{False Dilemma} (FD), \textit{Faulty Generalization} (FG), \textit{Intentional} (I), \textit{Logic} (L), and \textit{Relevance} (R).
An additional category, \textit{Miscellaneous}, was excluded as it was never relevant in our dataset.
Indeed, for each quip in each session, the classifier returned the probability that the text belongs to the aforementioned classes.
To determine whether a text should be considered fallacious or neutral, we introduced class-specific probability thresholds based on the empirical probability distributions across fallacy types. As distributions varied consistently between categories, heterogeneous threshold values were applied to mitigate over-classification as follows: Texts labeled as \textit{Equivocation} were retained if their score was $\geq$ 0.2; \textit{Logic}, \textit{Intentional}, and \textit{Extension} fallacies required a score $\geq$ 0.5; all other categories required a stricter threshold of $\geq$ 0.8. Predictions with probabilities below the corresponding threshold were conservatively classified as neutral, with no evidence of fallacies. As our texts were originally in Italian, they were translated into English prior to fallacy classification using Mistral Small 3.2. For the machine learning analyses, we used two aggregate features per session: the total number of fallacies detected in the user's and LLM's texts.

Finally, an additional pair of binary variables was extracted at the interaction level, namely ``user" and ``LLM credibility". For both, we evaluate whether the number of times a text is repeated is $\ge 3$. When this occurs in participants, it may indicate users who did not intend to complete the task; in LLMs, it indicates that the model was stuck in repetitive text generation. See \textit{SI, Feature Descriptions} for further details.

\subsection{Time-invariant Psychological Features}

The baseline cognitive profiles of the users was established from psychometric scores. Following a data validation of psychometric factors \cite{carrillo2026talk2ai}, the Trust in Large Language Models Index, denoted as \textit{TILLMI}, was operationalized as a single latent factor capturing user comfort, trust, and reliance on artificial assistants. A high score on this index indicated a disposition to treat generated outputs as dependable and to invest time in the interaction. The Need for Cognition construct was confirmed \cite{carrillo2026talk2ai} to divide into two distinct latent factors \cite{cacioppo1982need}. The seek factor, \textit{NFC-seek}, quantifies the motivational drive to engage in complex cognitive tasks and devise novel solutions. The diligence factor, \textit{NFC-diligence}, measures the behavioral persistence to sustain mental effort and complete rigorous tasks regardless of intrinsic enjoyment. The establishment of scalar invariance for these metrics ensured that factor structures and item intercepts remained equivalent over time. This structural stability permitted the valid integration of these latent scores throughout the longitudinal modeling. Details regarding specific factor structures, fit indices, and validation procedures are available in \cite{carrillo2026talk2ai}.

\subsection{Feedback Binning for Transitions}

Distributions across the four feedback metrics reveals that users concentrate around three anchors on the response slider: the minimum (0), the midpoint (50), and the maximum (100). Temporal dynamics, measured via absolute score deltas between sessions, peak around 0. This indicates high inertia, with a median absolute shift of roughly 10 points, $s_{th}=10$. This threshold, combined with the observed anchor points, was used to discretize continuous feedback scores into five categorical states: low anchor \textit{L} ($s_{th} \leq 10$), mid-low \textit{ML} ($10 < s_{th} < 40$), mid \textit{M} ($40 \leq s_{th} \leq 60$), mid-high  \textit{MH} ($60 < s_{th} < 90$), and high \textit{H} anchor ($s_{th} \geq 90$). 
Note, once again, that this binning aids the visualization of user trajectories via Sankey diagrams and Markov chains, and it is not used for predictions, for which we instead rely on the continuous feedback scores.

Transitions between feedback states were measured as follows: For each participant, we counted transitions from the first to the second session, from the second to the third, and from the third to the fourth, e.g. if a participant remained in the high category \textit{H} across all sessions, the self-loop counter for \textit{H} is incremented by 3. To ensure the robustness of the analysis, empirical transitions were compared against a null model in which each user's initial state was kept fixed while the remaining states were shuffled, preserving the overall dataset volume. Only transitions that were statistically significant according to a two-sided z-score ($\alpha=0.05$) were retained. Specifically, for each transition, $z=\frac{x-\mu}{\sigma}$ measures the deviation of an empirical transition count from the null expectation in units of standard deviations, where $x$ is the real count observed for a specified transition, e.g., $H \rightarrow H$, $\mu$ is the expected count under the null model, and $\sigma$ is the corresponding standard deviation.

\subsection{Feature Selection}
\begin{figure*}[t]
    \centering
    \includegraphics[width=16cm]{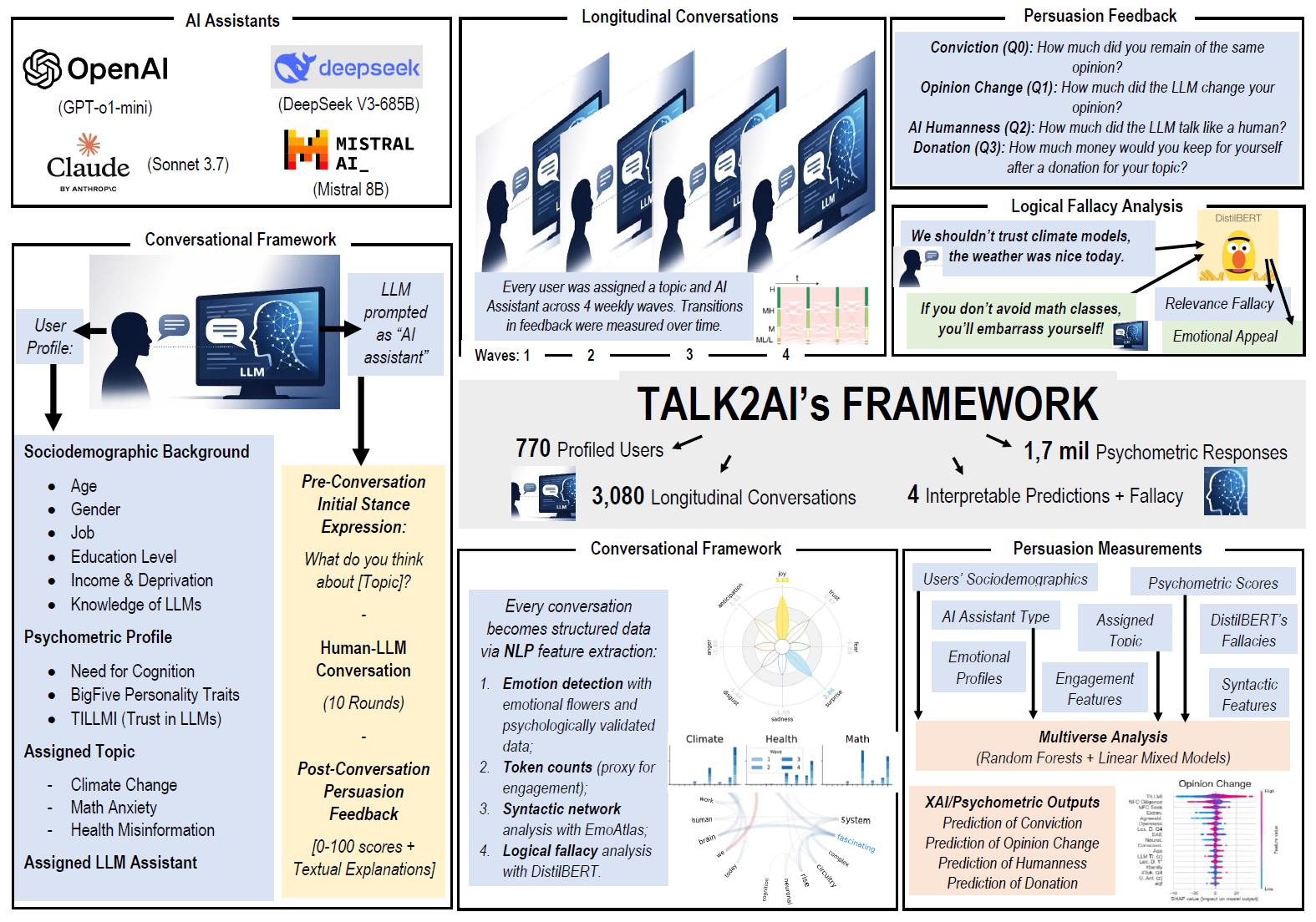}
    \caption{Infographics for Talk2AI's features.}
    \label{fig:1}
\end{figure*}

Each of the four feedback metrics ($Q0$-$Q3$) is represented as a continuous target variable and predicted by regression from an initial set of 58 features, cf. the infographics reported in Fig. \ref{fig:1}. To identify an optimal feature subset, we apply a Monte Carlo wrapper-based method. At each iteration, a Random Forest regressor is trained on the active feature set via randomized hyperparameter search with 5-fold cross-validation, optimizing $R^2$ score. Feature importances are estimated via mean absolute SHAP values, and the least important feature is removed from the active set and added to a removal pool. A stochastic reinsertion step is then performed: With reinsertion probability $p$, a feature is drawn from the removal pool with sampling probability $w$ proportional to its importance at the time of removal, thus reinserted into the active set. This mechanism enables exploration beyond the greedy elimination path, reducing the risk of converging to suboptimal feature subsets. The procedure terminates when performance drops more than $\varepsilon$ below the best observed $R^2$ or when only one feature remains. The final feature subset is taken as the union of the top-$k$ feature sets, to account for multiple near-equivalent solutions.We used $p = 0.65$, $\varepsilon = 0.25$, and $k=5$. The full algorithmic description and comparison with simpler recursive elimination strategies ($p=0$) are provided in \textit{SI, Feature Selection}.

% for each Feedback question, a different set of predictors .. as product of feature selection

% After feature selection, we implemented  two regression models: RFR and Linear mixed models (multiverse approach)
\subsection{Random forest regression} 

Random Forest regression, as implemented in \textit{scikit-learn 1.8}, was used to model the relationship between user, interaction, and linguistic features and the four feedback variables (Q0–Q3). Through model training, an optimal subset of features was identified through a wrapper-based Monte Carlo feature selection procedure (cf. Methods - Feature Selection) that iteratively removed the least informative variables based on SHAP importance while allowing stochastic reinsertion of previously discarded features (cf. SI for algorithmic details). The final feature set was defined as the union of the top-performing subsets ranked by cross-validated $R^2$. Models were evaluated using a 5-fold cross-validation, and predictive performance was assessed through $R^2$, root mean squared error (RMSE), and Pearson correlation between predicted and observed values. To interpret model predictions, SHAP (SHapley Additive exPlanations) values \cite{wang2024feature} were computed to estimate the contribution and direction of each feature’s effect on the model output, enabling interpretable comparison with the predictors identified in the mixed-effects analyses.

\subsection{Linear mixed model}

We fitted a Linear Mixed Model (LMM) \cite{bates2005fitting} for each of the four dependent feedback variables ($Y$: Conviction Stability, Opinion Change, Perceived Humanness, and Donation). The model specification was informed by the feature selection process, resulting in a high-dimensional set of predictors that controls for user characteristics, interaction content, and rhetorical strategies. The fixed effects were organized into two distinct categories based on the interaction term with the \textit{Time} variable:

\textit{Static Predictors (Main Effects Only).} These variables were modeled as stable baselines without a time interaction. This set includes experimental conditions (LLM Assistant, \texttt{llms}; Discussion Topic, \texttt{topic}); sociodemographics (Gender, Age, Education, \texttt{eqf}; Job Status, Household Size, Economic Evaluation, Financial Struggle, and Prior AI Knowledge); and core psychometrics (Trust in AI, \texttt{TILLMI}; Need for Cognition, \texttt{NFC\_seek}, \texttt{NFC\_diligence}).

\textit{Dynamic and Interacted Predictors (Main Effects $\times$ Time).} These features were interacted with the \textit{Time} variable to assess whether their predictive weight changed significantly across the four waves. This set includes interaction dynamics (repetition flags for user and assistant at 3 and 7 recurrences); rhetorical features (counts of logical fallacies committed by the user and the assistant); personality traits (the Big Five dimensions to test if the influence of personality moderates over repeated exposures); emotional profiling (normalized intensity $z$-scores for eight basic emotions for both user and assistant); and linguistic complexity (token counts and lexical diversity metrics for the opening turn and the qualitative feedback, \texttt{Q4}). 

\textit{Random Effects and Formal Specification.} To account for the non-independence of observations arising from the repeated-measures design, we included a random intercept for each participant (\texttt{userId}). The formal specification for the feedback variable $Y$ of user $i$ at time $t$ is expressed as:

\begin{equation}
    Y_{it} = \beta_0 + \beta_t t + \mathbf{X}_{i} \boldsymbol{\beta} + \mathbf{Z}_{it} \boldsymbol{\gamma} + t\mathbf{Z}_{it} \boldsymbol{\delta} + u_i + \epsilon_{it},
\end{equation}

where $\beta_0$ is the fixed intercept, $\beta_t$ captures the main effect of time $t$, $\mathbf{X}_{i}$ represents the vector of static predictors, $\mathbf{Z}_{it}$ represents the vector of dynamic predictors, $\boldsymbol{\beta}$, $\boldsymbol{\gamma}$, and $\boldsymbol{\delta}$ are the corresponding coefficient vectors, $u_i \sim \mathcal{N}(0, \sigma_u^2)$ is the random intercept for participant $i$, and $\epsilon_{it} \sim \mathcal{N}(0, \sigma_\epsilon^2)$ is the residual error term.

\section{Results}
% Narrative (suggested structure for results):
% this paper presents a double objective: First, analyze persuasion dynamics and second, identify the drivers of feedback.
% Our research examined the persuasive efficacy of LLMs through longitudinal user engagement across four conversational rounds. These interactions focused on climate change, mathematics anxiety, and health misinformation to assess shifts in participant belief systems. Analysis of the response dynamics across the temporal intervals indicates that individuals demonstrate cognitive anchoring, suggesting a high degree of belief stability despite recursive AI engagement.
% For the second objective, we performed an evaluation of response predictors with Random forest regressor and a linear mixed model. 
% The data science pipeline includes data analysis related to feature engineering, as fallacies or factor analysis, along with feature selection.
% We found that internal psychological factors serve as prevalent estimators of feedback scores. 

\subsection{Feedback Dynamics}

\begin{figure*}[t!]
\centering
\includegraphics[scale=0.44]{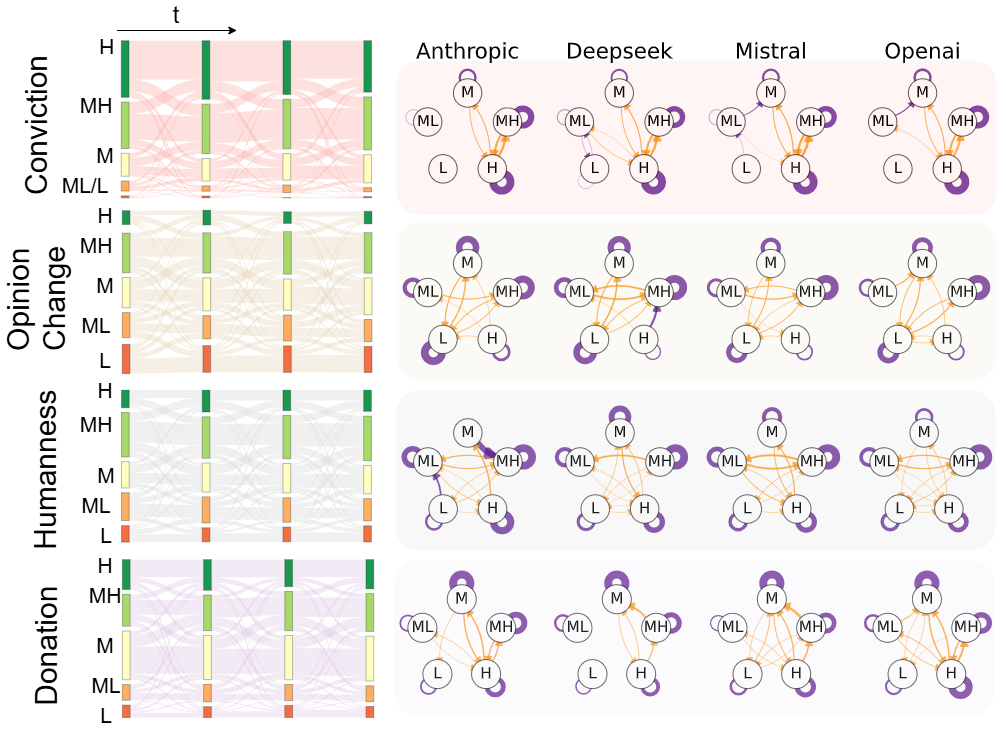}
\caption{Feedback transitions as Sankey flows across the four waves and Markov state transition patterns grouped by LLM interaction. In Markov chains, colored edges indicate the side of z-test: purple for more frequent-than-random transitions $z>1.96$; orange for less frequent-than-random transitions $z<-1.96$. Notice that "Donation" is short for "Self-donation".}
\label{fig:feedbacks}
\end{figure*}

Figure \ref{fig:feedbacks} shows trajectories of user feedback responses across the four waves (Sankey flows on the left), and transitions further grouped by LLMs (Markov chains on the right).
Continuous feedback responses were binned in five macro-categories, cf. \textit{Materials and Methods, Feedback binning and Transitions}.
Note that this discretization is used only for visualization purposes and not for the prediction tasks presented later in the work.
The Sankey diagrams describe a phenomenon in which participants remain largely anchored to their initial pole.
In addition, users starting at the extreme poles rarely migrate to the opposite pole in a single step, with medium-high ($MH$) and medium-low ($ML$) anchors functioning as bridges.
Markov chains convey this information as well: Overall, the consistent purple self-loops ($z>1.96$) indicate a higher tendency to remain in the same state compared to a null model in which states are shuffled except for the first one, cf. \textit{Materials and Methods, Feedback binning and Transitions}.
Notably, such edges connect adjacent states only, aligning with the gradual shifts evident in the Sankey diagrams.
Instead, orange edges ($z<-1.96$) are predominantly observed between between non-adjacent poles.
Note that orange edges occur significantly less frequently than expected under the null model. 
From Sankey flows, the high feedback pole ($H$) dominates in Conviction ($Q0$): Users starting highly convinced of initial arguments rarely drift to states of lower conviction.
Instead, Opinion Change ($Q1$) show a more heterogeneous distribution, rarely reaching the extreme pole ($H$), while the mid-high category ($MH$) exhibits a particularly dense flow.
This dynamics present is present also for Humanness ($Q2$).
Donation ($Q3$) shows a consolidation in the medium category ($M$), functioning also as a bridge of transition. 
Through the lens of Markov chains, orange edges are less frequent in Conviction than in the other feedback questions. Notably, for Anthropic, there are no transitions from or to the Low (L) and Medium-Low (M) poles, highlighting an even more limited influence in shifting users' opinions compared to Mistral or DeepSeek.

\subsection{Logical Fallacies in Human and LLM Argumentation}

\begin{figure*}[t!]
\centering
\includegraphics[scale=0.21]{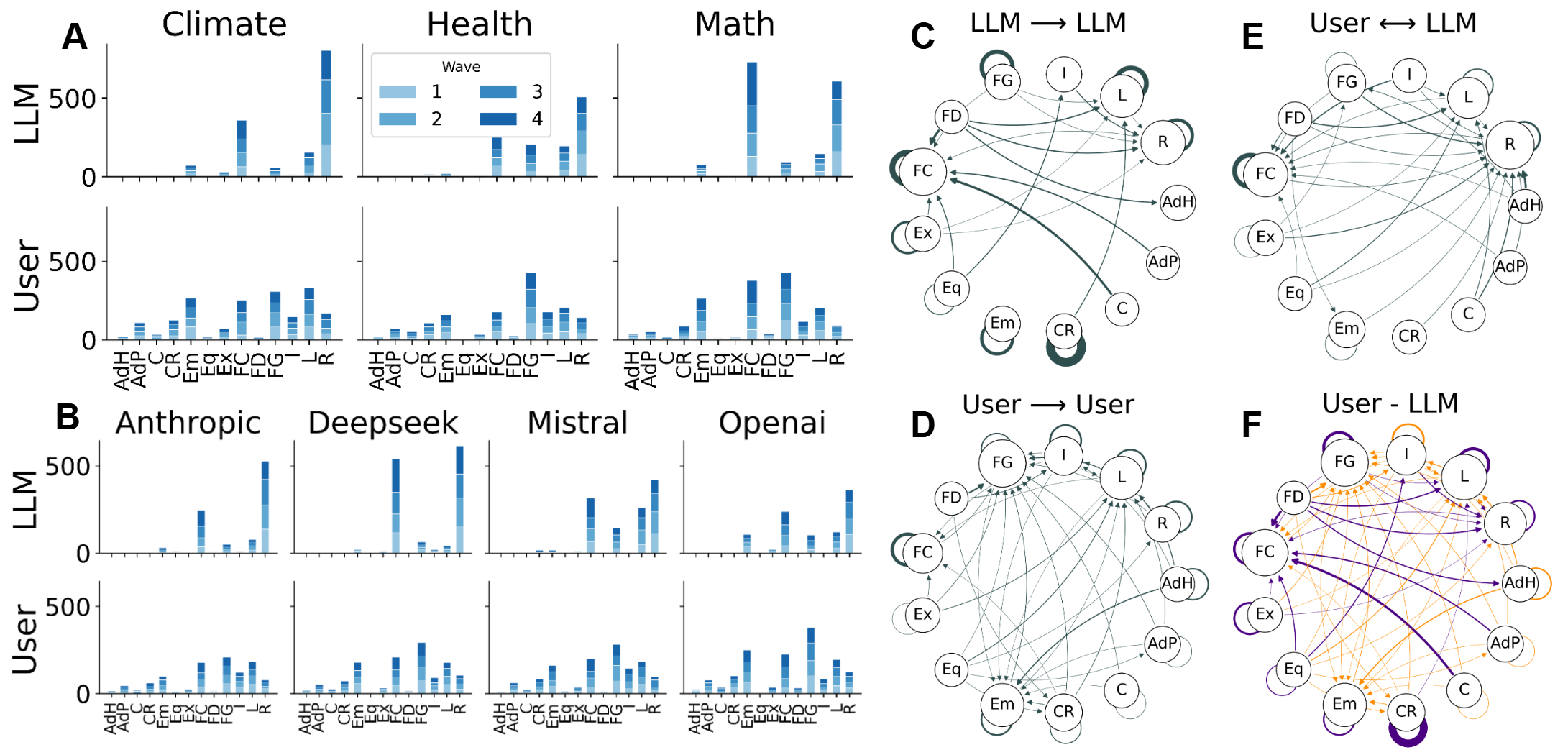}
\caption{(A-B) Frequency distribution of fallacies across quips, grouped by topic (A) and by LLM (B). \ (C-D) Fallacy transition patterns across the four aggregated waves showing how (C) LLMs and (D) users move between different fallacy types across quips. \ (E) Combined fallacy transitions (User $\rightarrow$ LLM $\rightarrow$ User $\rightarrow$  ...) across consecutive quips. (F) Differences in users' and LLMs' transition probabilities: Purple edges indicate higher probabilities for LLMs; orange edges, for users. Node sizes are proportional to in-degrees. Fallacies Legend --- AdH: Ad Hominem; AdP: Ad Populum; C: Credibility; CR: Circular Reasoning; Em: Appeal to Emotion; Eq: Equivocation; Ex: Extension; FC: False Causality; FD: False Dilemma; FG: Faulty Generalization; I: Intentional; L: Logic; R: Relevance.}
\label{fig:fallacies}
\end{figure*}

Logical fallacies \cite{jin2022logical} were identified, at the quip level, in 17\% of user texts and 14.5\% of LLM texts.
Fig. \ref{fig:fallacies}A and Fig. \ref{fig:fallacies}B highlight that LLM fallacies are concentrated in a few types, whereas users show greater heterogeneity. Across topics, AI assistants are less coherent in discussions on climate change (highest \textit{R} - Relevance Fallact) and they struggle to identify cause-and-effect relationships when discussing math anxiety (highest \textit{FC} - False Causality). DeepSeek shows the highest frequencies of these two fallacy types (Fig. \ref{fig:fallacies}B, first row).
Mistral produces logical fallacies (\textit{L}) more than other assistants, while OpenAI manifests the strongest tendency to rely on emotional persuasion (\textit{Em}).
Frequencies of emotional claims, however, are far lower than those produced by participants, as are other fallacies such as circular reasoning (\textit{CR}) and intentionality (\textit{I}).
Beyond counts, transitions reveal additional information about the patterns followed by LLMs (Fig. \ref{fig:fallacies}C), participants (Fig. \ref{fig:fallacies}D), and within the overall conversational flow (Fig. \ref{fig:fallacies}E).
AI assistants tend to follow more homogeneous patterns: Even starting from less frequent fallacies, e.g. credibility (\textit{C}) or equivocation (\textit{Eq}), they tend to converge on relevance, causality, and logical errors.
Moreover, assistants tend to remain on the same fallacy across quips more than humans (Fig. \ref{fig:fallacies}F, identified by the purple self-loops), except when dealing with intentionality, credibility, and arguments \textit{ad populum} (AdP) and \textit{ad hominem} (AdH).

\subsection{Random forest regression}

\begin{figure*}[t!]
\centering
\includegraphics[scale=0.38]{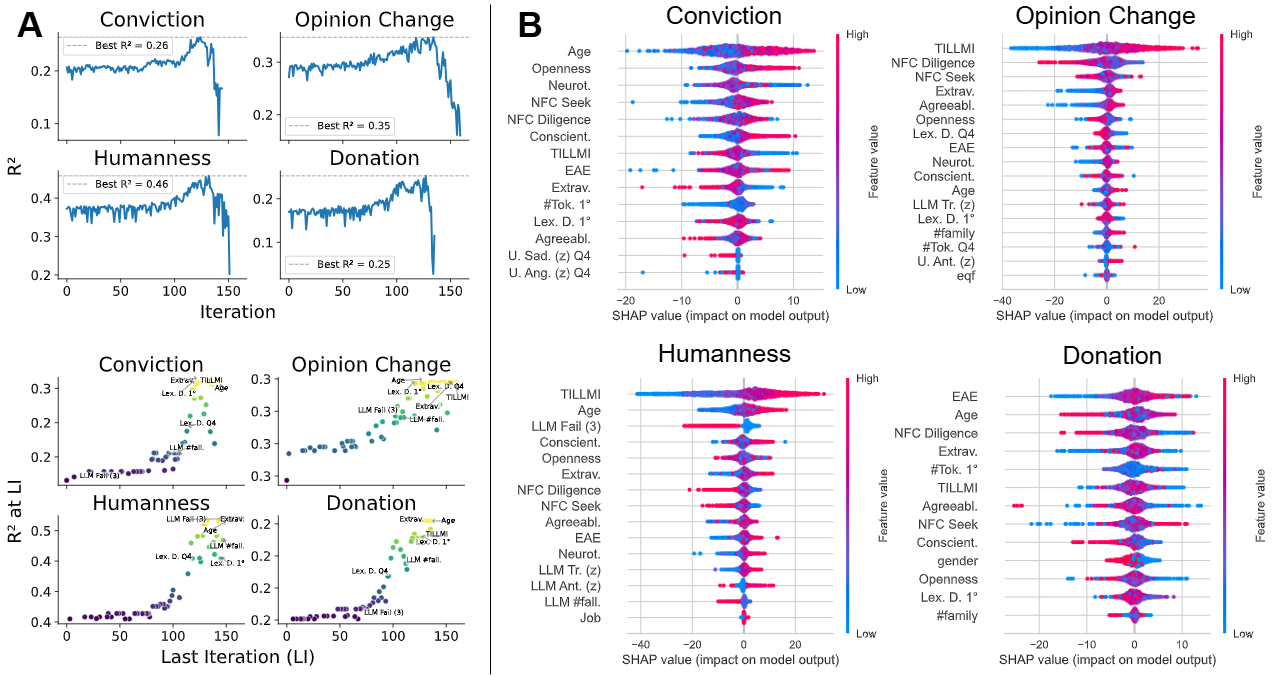}
\caption{(A) $R^2$ values of the feature selection method for identifying the best performing feature set (top); Scatter plot where each point is a feature, described by the last iteration $li$ where it occurs according to feature selection and its relative $R^2$ value at $li$ (bottom). (B) Feature importance according to beeswarms highligthing Shapley values. Feature space chosen as the union of the top-5 best performing feature subsets according to the wrapper feature selection method. Notice that "Donation" is short for self-donation in Q3.}
\label{fig:rf}
\end{figure*}

We trained a Random Forest (RF) regressor on the best-performing feature set including user- and interaction-level variables to predict $Q_0,Q_1,Q_2$ and $Q_3$.
To identify the optimal feature set, we implemented a wrapper-based feature selection procedure augmented with Monte Carlo sampling, cf. \textit{Materials and Methods, Feature Selection}.
Fig. \ref{fig:rf}A highlights multiple optimal peaks according to the $R^2$ scores (top).
Additionally, it shows the contribution of each feature at the stage when it is selected for removal (bottom).
This allows us to choose the union of the top-5 feature sets as the optimal feature subset, to account for multiple near-equivalent solutions.
Hence, Fig. \ref{fig:rf}B and Table \ref{tab:rf} present the results of a RF regressor trained on the feature set identified as described above.
Opinion Change ($R^2 = 0.34 \pm0.01$) and Humanness ($R^2 = 0.44 \pm0.01$) reach the highest predictive performance, whereas Conviction ($R^2 = 0.26 \pm0.03$) and Self-donation ($R^2 = 0.24 \pm0.04$) have lower maximum performance.
The SHAP beeswarm plots (Fig. \ref{fig:rf}B) highlight the importance and directional impact of each retained feature.
Overall, sociodemographic and psychometric features are the strongest predictors. 
Age emerges as a key factor for all feedbacks except Opinion Change, with older participants having a positive impact on the regression of Conviction and Humanness.
Psychometric variables emerge as the strongest predictors of Opinion Change, with trust in LLMs (TILLMI) showing the highest positive effect on both Opinion Change and Humanness.
Personality traits predict feedback differently \cite{guido2015italian}. Openness, for instance, proxied here by questions assessing an individual's creative disposition (cf. \textit{SI, Feature Descriptions}), shows a meaningful impact on Conviction, with higher values positively predicting this feedback. 
Among the features highlighted by the feature selection procedure, and despite their lower relative importance in the SHAP ranking, we also find variables related to initial engagement, proxied as the number of tokens in the first quip of the first wave (notably for Conviction) as well as emotional intensity measures extracted from the LLM-generated texts.

% features 

\begin{table}[t]
\caption{Random Forest performance on the best-performing feature set in terms of $R^2$, $RMSE$, and $\rho$ correlation.}
\label{tab:rf}
\centering
\small
\begin{tabular}{ccccccc}
\toprule
 & \multicolumn{2}{c}{R$^2$} & \multicolumn{2}{c}{RMSE} & \multicolumn{2}{c}{$\rho$} \\
\cmidrule(lr){2-3} \cmidrule(lr){4-5} \cmidrule(lr){6-7}
 & M & STD & M & STD & M & STD \\
\midrule
Q0 & 0.255 & 0.01 & 19.216 & 0.52 & 0.567 & 0.02 \\
Q1 & 0.339 & 0.03 & 25.369 & 0.62 & 0.578 & 0.03 \\
Q2 & 0.442 & 0.01 & 23.161 & 0.33 & 0.664 & 0.01 \\
Q3 & 0.240 & 0.04 & 25.098 & 0.80 & 0.472 & 0.04 \\
\bottomrule
\end{tabular}
\end{table}

\subsection{Linear mixed models}

Mixed effects models evaluate the influence of conversational dynamics across the four target variables. Fixed factors explain between 5.5\% of the variance for self-donation ($Q3$) and 21.4\% for perceived humanness ($Q2$). Including random factors increases the explained variance to 64.2\% for perceived humanness. 

\begin{table}[H]
\caption{Linear Mixed Model Performance}
\label{tab:model_performance}
\begin{tabular}{@{}lrrr@{}}
\toprule
Target Variable & Marginal $R^2$ & Conditional $R^2$ & RMSE \\ \midrule
Conviction Stability ($Q0$) & 0.111 & 0.442 & 0.152 \\
Self-Reported Opinion Change ($Q1$) & 0.174 & 0.554 & 0.185 \\
Perceived Humanness ($Q2$) & 0.214 & 0.642 & 0.161 \\
Self-donation ($Q3$) & 0.055 & 0.434 & 0.197 \\ \bottomrule
\end{tabular}%
\end{table}

Evaluating the dataset across artificial assistants reveals variations in predictability. Finally, assistant- and topic-specific multiverse results underscore that persuasiveness is not a fixed property of ``LLMs'' but an interaction between model style, user traits, and domain involvement. For example, Mistral maximizes fixed-effect explanatory power for humanness (Q2 marginal $R^2 = 0.316$), while Anthropic does so for opinion change (Q1 marginal $R^2 = 0.260$); topic-wise, Q2 is most predictable in math (marginal $R^2 = 0.303$), whereas Q3 remains modest even where highest (math marginal $R^2 = 0.151$). Within OpenAI interactions, $Q1$ correlates with disgust from bots and joy from users. Furthermore, the OpenAI model relies on establishing trust in bots and generating surprise in users to motivate actions of behavior ($Q3$). A topic-based analysis demonstrates a dependency of effectiveness for elements of conversation on the topic. For instance, the topic of mathematics maximizes the explained variance for humanness and shifts in opinions ($Q1$ and $Q2$) In the context of climate, the DeepSeek model drives shifts in opinions ($Q1$).

\begin{table}[t]
\caption{Linear Mixed Model Performance, separated by LLM}
\label{tab:llms_model_performance}
\centering
\begin{tabular}{@{}lrrr@{}}
\toprule
LLM & Marginal $R^2$ & Conditional $R^2$ & RMSE \\ \midrule
\multicolumn{4}{l}{\textbf{Conviction Stability (Q0)}} \\
OPENAI & 0.198 & 0.530 & 0.134 \\
ANTHROPIC & 0.217 & 0.441 & 0.145 \\
DEEPSEEK & 0.187 & 0.508 & 0.146 \\
MISTRAL & 0.185 & 0.519 & 0.141 \\
\multicolumn{4}{l}{\textbf{Self-Reported Opinion Change (Q1)}} \\
OPENAI & 0.254 & 0.651 & 0.164 \\
ANTHROPIC & 0.260 & 0.557 & 0.173 \\
DEEPSEEK & 0.227 & 0.526 & 0.180 \\
MISTRAL & 0.255 & 0.629 & 0.167 \\
\multicolumn{4}{l}{\textbf{Perceived Humanness (Q2)}} \\
OPENAI & 0.226 & 0.692 & 0.140 \\
ANTHROPIC & 0.265 & 0.647 & 0.147 \\
DEEPSEEK & 0.266 & 0.618 & 0.157 \\
MISTRAL & 0.316 & 0.695 & 0.151 \\
\multicolumn{4}{l}{\textbf{Self-Donation (Q3)}} \\
OPENAI & 0.146 & 0.481 & 0.183 \\
ANTHROPIC & 0.187 & 0.586 & 0.158 \\
DEEPSEEK & 0.147 & 0.447 & 0.199 \\
MISTRAL & 0.148 & 0.529 & 0.184 \\ \bottomrule
\end{tabular}%
\end{table}

% include paragraph on performance by topic

Interestingly, the contextual domain of the interaction further dictates the explanatory power of the regression models. For conviction stability, fixed effects explain the maximum variance in health discussions. Total explained variance including individual differences peaks in the climate context. Predictions of opinion change demonstrate comparable fixed-effect performance for health and mathematics discussions. Climate discussions yield the highest total variance explained for this metric. Perceived humanness achieves the highest predictive accuracy among all categories. Mathematics discussions maximize both fixed and total explained variance for humanness evaluations. Conversely, self-donation/behavioral persuasion (Q3) exhibits the lowest predictability across all subjects. Mathematics interactions provide the highest fixed-effect variance for behavioral outcomes. Hence, the contextual domain of the interaction dictates the explanatory power of the predictive models.

\begin{table}[t]
\caption{Linear Mixed Model Performance, separated by topic}
\label{tab:topic_model_performance}
\centering
\begin{tabular}{@{}lrrr@{}}
\toprule
LLM & Marginal $R^2$ & Conditional $R^2$ & RMSE \\ \midrule
\multicolumn{4}{l}{\textbf{Conviction Stability (Q0)}} \\
CLIMATE & 0.165 & 0.529 & 0.139 \\
HEALTH & 0.227 & 0.468 & 0.144 \\
MATH & 0.152 & 0.443 & 0.152 \\
\multicolumn{4}{l}{\textbf{Self-Reported Opinion Change (Q1)}} \\
CLIMATE & 0.184 & 0.595 & 0.176 \\
HEALTH & 0.249 & 0.576 & 0.179 \\
MATH & 0.251 & 0.572 & 0.175 \\
\multicolumn{4}{l}{\textbf{Perceived Humanness (Q2)}} \\
CLIMATE & 0.241 & 0.652 & 0.157 \\
HEALTH & 0.231 & 0.654 & 0.156 \\
MATH & 0.303 & 0.688 & 0.148 \\
\multicolumn{4}{l}{\textbf{Self-donation (Q3)}} \\
CLIMATE & 0.109 & 0.491 & 0.191 \\
HEALTH & 0.122 & 0.431 & 0.192 \\
MATH & 0.151 & 0.483 & 0.183 \\ \bottomrule
\end{tabular}%

\end{table}

\subsection{Multiverse analysis} The multiverse analysis further shows that machine-learning and mixed-effects approaches converge on the same explanatory structure of persuasion, providing complementary but conceptually aligned evidence. Random Forest models with SHAP-based feature attribution identify humanness and opinion change as the most predictable outcomes ($R^2 \approx 0.44$ for Q2 and $R^2 \approx 0.34$ for Q1), with conviction ($R^2 \approx 0.26$) and donation ($R^2 \approx 0.24$) substantially less predictable from the feature space. Mixed-effects models replicate this ordering: fixed conversational and contextual factors explain the most variance for perceived humanness (marginal $R^2 \approx 0.21$) and less for opinion change ($R^2 \approx 0.17$), conviction ($R^2 \approx 0.11$), and self-donation ($R^2 \approx 0.06$), while including participant-level random effects raises the explained variance markedly (conditional $R^2 \approx 0.64$ for humanness and $R^2 \approx 0.55$ for opinion change). The convergence extends beyond aggregate fit statistics: The predictors highlighted by SHAP—especially trust in LLMs (TILLMI \cite{de2025measuring}), age, and personality traits—also emerge as significant fixed effects in the mixed models (e.g., TILLMI strongly predicting both Q1 and Q2), while emotional signals and fallacy-related conversational features appear as significant interaction-level predictors in the longitudinal regressions (see SI Tables of significant predictors). Methodologically, this alignment illustrates the value of a multiverse strategy that integrates interpretable machine learning with hierarchical statistical modeling: ML captures nonlinear combinations and feature salience in a high-dimensional space, whereas mixed models validate the directionality and temporal structure of those relationships while accounting for repeated observations and stable individual differences. The fact that both approaches independently converge on the same hierarchy of outcomes and predictors strengthens the robustness of the Talk2AI framework and supports the interpretation that persuasion in conversational AI is driven by stable user traits and trust-related heuristics more than by transient conversational cues alone.

\section{Discussion}

Talk2AI is designed to test whether conversational LLMs can persuade people over time. Its central result is a layered pattern of LLM-based influence over time and across specific categories of users: Across four weekly waves, participants remained strongly anchored to their initial conviction poles, yet showed greater movement in self-reported opinion change and perceived humanness. This distinction is theoretically important because it separates temporary impressions of influence from more durable changes in conviction, and it situates conversational AI within both dual-process models of persuasion and social-response theories of human--machine interaction \cite{chaiken1989heuristic,petty2009elaboration,gambino2020building,xu2022deep}.

Our key findings are \emph{longitudinal inertia} and \emph{psychological susceptibility}: While many participants predominantly remained in nearby response states across waves (inertia), opinion changes were reported in specific categories of users, with well-identified personality traits, knowledge search cognitive traits, trust in AI heuristics and socio-demographic features (susceptibility). Importantly, these two elements do not exclude each other, considering that inertia does not imply an absence of LLMs' influence. Recent work on resistance to persuasion shows that post-message \emph{attitude certainty} and downstream intentions can vary as a function of how recipients resist an appeal, even when the underlying attitude itself does not move in parallel \cite{blankenship2023certainty}. Read in this light, Talk2AI's separation between conviction stability (Q0) and self-reported opinion change (Q1) is theoretically meaningful rather than contradictory: Users may recognize the plausibility or force of an LLM-generated argument while still preserving a stable attitudinal anchor. Such dissociation is also visible in the predictive hierarchy of outcomes. In explainable AI models, humanness is the most learnable target ($R^2 = 0.44 \pm 0.01$), followed by opinion change, conviction, and donation (cf. Results -- Multiverse analysis). The key implication is that conversational AI may be especially effective at shifting social appraisal \cite{breum2024persuasive} and metacognitive judgments \cite{shaw2026thinking} of being influenced before it produces durable belief revision or costly action. This interpretation is consistent with persuasion research showing that certainty, agreement, and behavioral readiness are related but non-identical components of response to an appeal \cite{blankenship2023certainty,shaw2026thinking,rogiers2024persuasion}. In practical terms, Q1 (Opinion Change) should therefore be interpreted less as a direct proxy for stable belief change than as a participant's subjective sense of having been influenced during the interaction. Importantly, higher Openness corresponded here to more artistic mindsets, following the encoding of personality traits in the Big Five Inventory-10 implementation \cite{guido2015italian}. These artistic mindsets can be more elaborative and cognitively engaged \cite{deyoung2014openness}, entertaining alternative arguments while still keeping a clearer, better-structured sense of their prior position or conviction. This reconciles with our XAI pattern of higher Openness corresponding to higher Conviction.

Humanness emerges as the most structured XAI pattern, and recent human--AI research clarifies why. Meta-analytic evidence on text-based conversational agents shows that human-like social cues reliably increase users' perceptions, affect, rapport, and trust, whereas behavioral consequences are markedly smaller \cite{klein2025socialcues}. This pattern closely mirrors the ordering recovered in Talk2AI, where Q2 is substantially more predictable than Q3. It also aligns with longitudinal work on AI advice showing that trust and reliance become path-dependent across repeated interactions: Prior trust shapes later trust, and prior reliance shapes later reliance \cite{kahr2024trust}. Seen from this perspective, perceived humanness is not a superficial stylistic judgment but a psychologically meaningful cue that calibrates whether the system is treated as a socially credible partner. This also helps explain why trust in LLMs (TILLMI, \cite{de2025measuring}) and other stable user-level traits emerge so prominently in the prediction of Q2 and Q1. 

Donation decisions are comparatively weakly explained by our models. This is consistent with meta-analytic evidence that AI communicators are roughly as persuasive as human communicators overall, but comparatively weaker at shifting behavioral intentions than perceptions or attitudes \cite{huang2023persuasive}. The dissociation between a strong social-perceptual pathway (Q2) and a weaker behavioral pathway (Q3) therefore argues against an overgeneralized claim that LLMs straightforwardly drive behavior in this paradigm, while still highlighting the upstream mechanisms (e.g. trust \cite{de2025measuring,branda2026comfort}) through which risk could accumulate in repeated real-world deployments.

Another important result concerns fallacious arguments. In our data, fallacious reasoning is not rare \cite{jin2022logical}, occurring in 17\% of user quips and 14.5\% of LLM quips. This importantly counters the stereotype of LLMs as superior statistical systems or perfect oracles, which has been hypothesized being at the base of cognitive offloading \cite{risko2016cognitive}, sovereignity \cite{branda2026comfort} or surrender \cite{shaw2026thinking} phenomena - where humans overtrust LLMs and accept LLM-based responses without critical thinking \cite{argyle2025testing}. In our case, also LLMs can fall into logical fallacies, with rates similar to humans' but with rather different structural properties: Humans show broader heterogeneity of fallacy types, whereas LLM fallacies concentrate into fewer categories and display more homogeneous transition patterns. Recent cognitive work helps interpret this asymmetry. Marin and colleagues \cite{marin2024poorarguments} showed that people are more accepting of poorly justified arguments when those arguments align with their prior attitudes, whereas scientific reasoning ability and active open-mindedness improve fallacy detection. This suggests that the heterogeneity of human fallacies in Talk2AI may partly reflect topic-sensitive, attitude-laden everyday reasoning, whereas the more homogeneous LLM profile may reflect recurrent model-level shortcuts for producing fluent but structurally weak justifications.

Taken together, Talk2AI provides strong longitudinal, empirical evidence that LLMs most reliably shift social-perceptual and credibility-related appraisals, which can alter subjective impressions of influence. Instead, durable conviction change and costly action remain bounded by attitudinal anchoring and stable individual differences. The mixed-model variation across topics and assistants is therefore better interpreted through persuasion theory's moderator logic than through a simple strong-vs.-weak persuasion dichotomy.

\subsection{Limitations and Future Research}
Several limitations should be considered when interpreting these findings. First, the study relies on self-reported persuasion metrics (Q0–Q2) and a hypothetical allocation task (Q3), which may capture perceived influence rather than durable attitude change or real-world behavior. Future work could address this limitation by incorporating incentive-compatible behavioral outcomes, delayed follow-up measures, or real-world decision tasks that assess persistence and external validity of conversational influence. Second, although the longitudinal design improves on one-shot persuasion experiments, the interaction structure was standardized (ten turns per session, weekly intervals), which may constrain natural conversational dynamics. Subsequent studies could examine more ecologically valid settings, including open-ended interactions, longer exposure periods, and adaptive dialogue structures that better approximate everyday AI use. Third, linguistic and fallacy detection relied on automated classification pipelines, which, despite enabling large-scale analysis, may overlook contextual nuances or culturally specific rhetorical strategies. Combining computational detection with human annotation or hybrid validation frameworks would strengthen interpretability. Fourth, the participant sample, while sizeable (N=770), originates from a specific cultural and linguistic context, potentially limiting generalizability. Cross-cultural replications and multilingual conversational datasets would help determine whether the observed persuasion pathways generalize across sociotechnical environments. Finally, the study focuses on four LLM systems available at the time of data collection; rapid model evolution implies that persuasive affordances may shift as architectures and alignment strategies change. Continuous benchmarking frameworks such as Talk2AI will therefore be essential for monitoring the evolving persuasive dynamics of conversational AI systems.

\section*{Data Availability}
All the data and code developed within the Talk2AI framework is available on the following GitHub: https://github.com/MassimoStel/Talk2AI

\section*{Acknowledgement}
The authors acknowledge support from the following grants: Call for Research Grant 2023 funded by University of Trento (ID: PS 22\_27, A.C., E.T., G.A.V. and M.S.); CALCOLO project funded by Fondazione VRT (M.S.); FIS project funded by Ministero dell'Università e della Ricerca, D.D.N. 23178, 10/12/2024, BANDO FIS2, ID: FIS-2023-02086 (S.C., A.A.A., M.S.).

 \bibliographystyle{elsarticle-harv} 
 \bibliography{bibliography}

\end{document}